\documentclass[letterpaper, 10 pt, conference]{ieeeconf}  % Comment this line out if you need a4paper

\IEEEoverridecommandlockouts                              % This command is only needed if 
\usepackage{amsmath}    % 核心：支持分式、矩阵、多行公式、上下标等
\usepackage{amssymb}    % 扩展：提供更多数学符号（如ℝ、∅、∑、∏等）
\usepackage{amsfonts}   % 辅助：提供数学字体（如黑体、花体）
\usepackage{mathtools}  % 进阶：amsmath的增强版，兼容IEEE，支持更灵活的公式排版
\usepackage{upgreek}    % 可选：非斜体希腊字母（如$\upalpha$）
\usepackage{graphicx} % 导入图片必备包
\usepackage{float}    % 可选：强制图片位置（如H）
\usepackage{booktabs}
\usepackage{multirow}
\usepackage{cite}
\usepackage{xcolor}
\usepackage{booktabs} % 用于 \toprule, \midrule, \bottomrule, \cmidrule
\usepackage{array}    % 用于 
\usepackage{bbding}
\usepackage{multirow}
\usepackage{graphicx}
\title{\LARGE \bf
Multiple cyclicity and Wavelet Decomposition with Channel Correlation for Long-term Time Series Forecasting
}

\author{Bin Wang$^{1}$, Heming Yang$^{1}$ and Jinfang Sheng$^{1*}$% <-this % stops a space
\thanks{$^{1}$Bin Wang, Heming Yang and Jinfang Sheng are all with 
School of Computer Science and Engineering, Central South University, HuNan, China. \{ wb\_csut, 244712142, jfsheng\}@csu.edu.cn}%
\thanks{*Corresponding author: Jinfang Sheng}%
}

\begin{document}

\maketitle
\thispagestyle{empty}
\pagestyle{empty}

%%%%%%%%%%%%%%%%%%%%%%%%%%%%%%%%%%%%%%%%%%%%%%%%%%%%%%%%%%%%%%%%%%%%%%%%%%%%%%%%
\begin{abstract}
Cyclicity and trend are important components of time series data and many studies based on cyclicity and trend have achieved good results in long-term time series forecasting. However, we believe that current work neglects the influence of real-world inter-channel correlations in time series data which leads to suboptimal predictions. Furthermore, these models rely on complex designs to capture diverse information so that resulting in low computational efficiency. To address this challenge, we propose McWC, a long-term time series forecasting model that separately models the cyclicity, trend, and inter-channel correlations. Specifically, McWC first decouples cyclical information from data using a multi-layer cyclicity construction module. Then, it extracts inter-channel correlations using multi-layer perceptron. Next, it models and fuses the multi-layer high-frequency and low-frequency information from data using a multi-level wavelet decomposition module. Finally, it aggregates the results of different components to obtain the output. Simultaneously, we decouple intra-channel autocorrelations by calculating a loss function in the frequency domain. Experiments on six real-world datasets demonstrate that McWC achieves state-of-the-art performance, exhibiting excellent computational efficiency and historical information extraction capabilities.
\end{abstract}

%%%%%%%%%%%%%%%%%%%%%%%%%%%%%%%%%%%%%%%%%%%%%%%%%%%%%%%%%%%%%%%%%%%%%%%%%%%%%%%%
\section{INTRODUCTION}

Long-term time series forecasting(LTSF) has been an important part in some key areas, such as financial forecasting and analysis\cite{zhang_financial_2023}, IoT device information prediction\cite{zhang2024sageformer}, urban traffic forecast\cite{hu_decomposition_2025,hadry_telling_2025}, and energy and resource allocation forecasting\cite{qin2026forecasting}. However, because long-term time series are real-world sequential data with complex and non-stationary characteristics, it poses significant challenges for researchers to capture meaningful dependencies from extensive historical data.

With the development of deep learning, various architectures have been proposed for time series modeling. CNN-based models\cite{lecun2002gradient} capture local temporal and channel dependencies via fixed kernels, but perform poorly in modeling global dependencies. Transformer-based models\cite{vaswani_attention_2017} use self-attention to extract intrinsic relations and achieve accurate long-term forecasting, yet they tend to focus on high frequency information while ignoring useful low-frequency components, thus limiting representation diversity. Linear-based models decompose time series into basic components and model them with simple linear operations. Methods such as trend-cycle decomposition and Fourier-based downsampling have achieved competitive results even with low computation and parameters. Therefore, we argue that properly modeling dependencies in time series is the key to high performance long-term forecasting.

We propose that time series data is composed of cyclical patterns, trend, and inter-channel corrections components. Influenced by the current channel-independent approach, many models ignore inter-channel corrections and still achieve good results by only modeling cyclical patterns and trend. However, we argue that since inter-channel corrections objectively exist, modeling them can improve the model's forecasting performance to some extent. Meanwhile, most existing models either directly capture temporal relationships in the time series data or construct simple cyclical patterns which prevents temporal relationships from being well represented. Therefore, we believe that constructing multiple cyclical patterns based on the inherent periods of the time series data can effectively model the cyclical patterns in the data.

In summary, we propose McWC, which uses the Multi-cycle Construction Block (McB) to extract cyclical patterns, then captures inter-channel correlations through the Channel-Correction Extraction Block (CEB) and finally extracts trend through our designed Multi-level Wavelet Decomposition Block (MWB). Through our design, the main components of the time series data are separately extracted and modeled by three interrelated yet independently handled parts, allowing each component to be fully utilized by the model, thereby achieving better forecasting performance. Our tests show that McWC achieves excellent performance on a wide range of long-term forecasting tasks. Our contributions in this paper can be summarized as follows:

\begin{enumerate}
  \itemsep=0pt
  \item We proposed McB based on the idea of explicitly constructing cyclical patterns using prior periodic information to capture periodic patterns of different frequencies within the time series data. And we proposed MWB module which uses multi-level wavelet decomposition to extract trend and models frequency-domain information at different scales, eliminating noise interference while capturing trend changes effectively.
  \item We proposed CEB, which models dynamic interactions between different scales, handling inter-channel corrections at various scales and enhancing the model's ability to process multi-scale information.
  \item Our designed McWC uses three interrelated but separately modeled components for time series forecasting. McWC demonstrates outstanding performance with remarkable efficiency across multiple long-term time series forecasting tasks and datasets.
  \end{enumerate}

\section{Related Work}
The primary process of LTSF involves analyzing and extracting features from past time series data to predict future values. Therefore, how to properly decompose the data into different components and extract inter-channel corrections to achieve accurate future predictions has always been a core concern for researchers.

In terms of decomposition, FEDformer\cite{zhou2022fedformer} employs a multi-kernel moving average to enhance its decomposition capability. As a Linear-based model, DLinear\cite{zeng_are_2023} employs a moving average method to decompose time series data into seasonal and trend components and uses two separate single-layer linear networks to model these components for prediction, achieving better results than Autoformer\cite{wu_autoformer_2021} which uses a Transformer architecture. Guided by cyclicity and trend decomposition, MICN\cite{wang_micn_2023} decomposes data into Seasonal and Trend-Cyclical terms and combines correlations between local features to achieve high-quality forecasting. TimesNet\cite{wu_timesnet_2023} employs Fourier transform to detect temporal periodicities, converts 1D sequences into 2D tensors and leverages Inception CNN to capture intra-period and inter-period variations. It then achieves high-precision prediction through temporal unfolding and adaptive aggregation. PatchTST\cite{nie_time_2023} decomposes continuous time series data into individual patches and uses a Transformer model for forecasting with good results. TimeMixer\cite{wang_timemixer_2024}, which adopts a multi-level decomposition pattern obtains time series at different scales through downsampling and decomposes them into seasonal and trend series at each scale to model the different components. TimeKAN\cite{huang_timekan_2025} utilizes the FreTS\cite{yi_frequency-domain_2023} time-frequency conversion concept, decomposes frequency-domain information at different scales and uses the modeling capability of KAN networks to extract trend. WPMixer\cite{murad2025wpmixer} uses wavelet transforms to model the high-frequency and low-frequency components of the time series data separately.

For inter-channel dependency extraction: despite strong performance of channel-independent strategies, researchers explore encoders capturing both intra and inter-channel interactions. Crossformer\cite{liang_crossformer_2024} uses two-stage attention to hierarchically process temporal and variable dimensions, enhancing model comprehension. iTransformer\cite{liu_itransformer_2024} inverts time series and uses self-attention to capture inter-channel dependencies. CARD \cite{wangcard} introduces channel-aligned attention to capture temporal and multi-channel dynamic dependencies, plus a token mixing module for multi-resolution tokens and multi-scale knowledge utilization. SDE\cite{weng2025sde} leverages state space models for LTSF, capturing temporal dynamics and inter-channel dependencies simultaneously.

Different from the previous methods, our proposed McWC combines the strengths of cyclicity-trend decomposition and inter-channel corrections extraction and leverages the powerful capability of multi-scale analysis in extracting trend to enhance the model's forecasting accuracy.

\section{McWC}
\begin{figure*}[!t]
	\centering
	\includegraphics[width=0.85\textwidth]{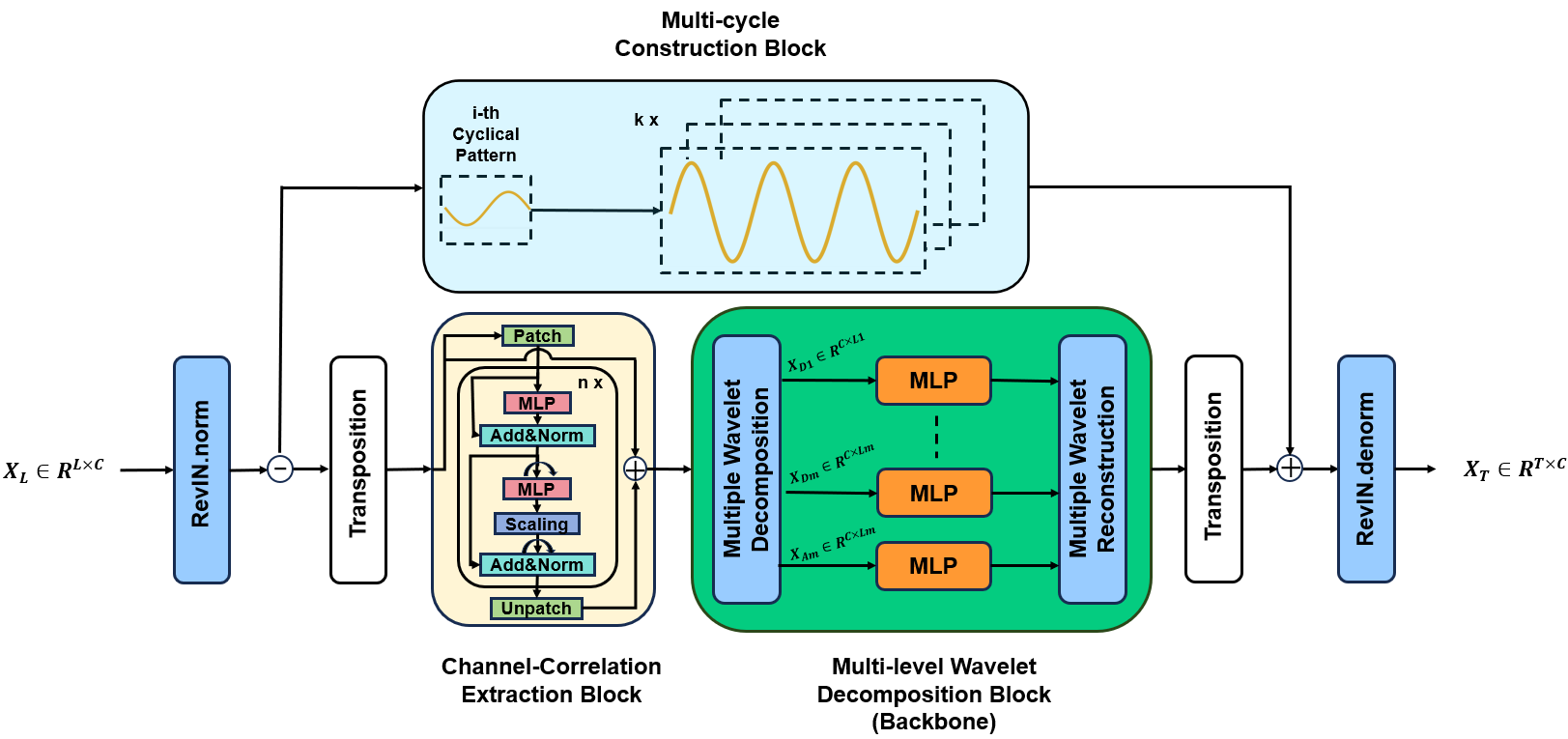}
	\caption{The architecture of McWC.}
	\label{FIG:1}
\end{figure*}
\subsection{OverView}\label{subsec1}
The overall architecture of McWC is shown in Figure 1. It consists of three modules: the McB module for constructing cyclical patterns, the CEB module for extracting inter-channel correlations, and the backbone forecasting module MWB. Through these modules, McWC decomposes, learns, and integrates long-term time series data, thereby achieving high-quality forecasting performance.

\subsection{Normalization \& Multi-cycle Construction Block}\label{subsec3}

First, we use \textbf{RevIN} method to normalize time series data. Then we identify the top-k period lengths $p=\left\{p_1,p_2\ldots p_k\right\}$ from data using prior knowledge, where pi denotes the period of the i-th highest weight cyclical pattern. For each period $p_i$, we construct a globally shared, learnable matrix $M_i\in R^{p_i\times C}$ initialized to zero. By iteratively copying $M_i$, we obtain a periodic component $Cyclist_i\in R^{L\times C}$ with the same length as the input sequence and subtract the periodic component from the normalized data. All periodic components and the prediction backbone are jointly optimized via backpropagation to update the learnable matrices, enabling accurate modeling of cyclical patterns. This procedure is formalized in formula(1):
\begin{eqnarray}
  X=RevIN(X)-\sum{Cyclist_i}
\end{eqnarray}
This allows subsequent models to ignore periodic patterns and focus on building trend and inter-channel corrections and improve prediction quality.

\subsection{Channel-Correction Extraction Block}\label{subsec4}
In CEB, we first split the input into patches based on the time series locality principle. For each patch, an MLP aggregates information to efficiently process both global and local details. We then transpose data dimensions and apply another MLP to blend cross-channel information, reverse the patch-splitting operation to restore the original dimensions, and add the result to the original input as supplementary information for subsequent steps. The entire block process is represented by Formula (2) and Formula (3):

\begin{eqnarray}
  X_{Patch}=Patch\left(X\right)+MLP\left(Patch\left(X\right)\right)
\end{eqnarray}

\begin{eqnarray}
  X=X+\alpha\cdot \left( X_{Patch}+\beta\cdot MLP\left(X_{Patch}^T\right)^T\right)
\end{eqnarray}

Here, $\alpha$ is a trainable model parameter, while $\beta$ is a manually defined scaling weight fixed during training. This design lets inter-channel dependencies in sequential data either be learned automatically from historical data or explicitly regulated via prior knowledge. Consequently, the model prediction remains robust when inter-channel correlations are weak or absent, thus enabling effective modeling of inter-channel dependencies.

\subsection{Backbone(MWB)}\label{subsec5}
In the MWB, we first use multi-level wavelet decomposition to split the input into a low-frequency component and multiple high-frequency components. We then assign an independent MLP to each decomposed component; these MLPs operate within their respective frequency domains to independently map historical data to forecasted time series. Finally, the model reconstructs the complete time series via inverse wavelet decomposition of all high and low frequency forecast results. This process is represented by Formula (4) and Formula (5):

\begin{eqnarray}
  \left[X_A,X_{D_1}\ldots X_{D_m}\right]=WaveletDecomp\left(X\right)
\end{eqnarray}

\begin{align}
X &= \text{WaveletReconstruction}\bigl( \bigl[ \text{MLP}(X_A), \text{MLP}(X_{D_1}), \notag \\
  &\quad \dots, \text{MLP}(X_{D_m}) \bigr] \bigr)
\end{align}

Here, $X_A$ is the low-frequency signal obtained from the multilevel wavelet decomposition of the input data, and $X_{D_1}\ldots X_{D_m}$ are the high-frequency signals obtained at each level of the decomposition. In this way, the multilevel wavelet decomposition module achieves the extraction of trend variations from the sequential data.

\subsection{Component Fusion}\label{subsec6}
Before producing the final output, we need to fuse the components and apply the inverse normalization of RevIN to align the distributional characteristics of the result with the original sequence as formula(6) :

\begin{eqnarray}
  X=iRevIN(X+\sum{Cyclist_i^\prime})
\end{eqnarray}

In this context, $Cyclist_i^\prime$ denotes the cyclical pattern for the k-th cycle and Y is the final forecast generated by the model.

\subsection{Train Loss}\label{subsec7}
To solve the problem that MSE loss struggles to handle intra-channel autocorrelation, we introduce FreDF\cite{wang_fredf_2024} as the loss of our model to better address autocorrelation in sequence data.
\begin{align}
\text{Loss} &= \alpha \times \bigl| \text{FFT}(\text{pred}) - \text{FFT}(\text{real}) \bigr|_1 \notag \\
           &\quad + (1-\alpha) \times \text{MSE}
\end{align}

Here, $\alpha$ is the weight for the frequency-domain loss, $FFT()$ denotes the Fast Fourier Transform operation, $pred$ represents the prediction and $real$ represents the ground truth value. This enables the model to better handle autocorrelation issues within channels by modifying the model's loss function by transforming time-domain data into the frequency-domain.

\section{Experiments}
\begin{table*}[!t]
  \centering
  \caption{Performance comparison of different time series forecasting models on benchmark datasets. The best results are highlighted in {\color[HTML]{FF0000} \textbf{bold red}} and the second-best results are shown in \textbf{bold black}. }
  \label{tab:comparison}
  \resizebox{\textwidth}{!}{%
    \setlength{\tabcolsep}{2pt}
    \begin{tabular}{cc|cc|cc|cc|cc|cc|cc|cc|cc}
\midrule
\multicolumn{2}{c|}{}                                & \multicolumn{2}{c|}{McWC}                                                     & \multicolumn{2}{c|}{WPMixer}                                                  & \multicolumn{2}{c|}{SDE}                               & \multicolumn{2}{c|}{TimeMixer} & \multicolumn{2}{c|}{iTransformer} & \multicolumn{2}{c|}{FreTS}                    & \multicolumn{2}{c|}{PatchTST} & \multicolumn{2}{c}{Dlinear} \\ \cmidrule{3-18} 
\multicolumn{2}{c|}{\multirow{-2}{*}{Models}}        & \multicolumn{2}{c|}{Ours}                                                     & \multicolumn{2}{c|}{2025}                                                     & \multicolumn{2}{c|}{2025}                              & \multicolumn{2}{c|}{2024}      & \multicolumn{2}{c|}{2024}         & \multicolumn{2}{c|}{2024}                     & \multicolumn{2}{c|}{2023}     & \multicolumn{2}{c}{2023}    \\ \midrule
\multicolumn{2}{c|}{Metric}                          & MSE                                   & MAE                                   & MSE                                   & MAE                                   & MSE                                   & MAE            & MSE                & MAE       & MSE         & MAE                 & MSE                                   & MAE   & MSE                & MAE      & MSE          & MAE          \\ \midrule
\multicolumn{1}{c|}{}                          & 96  & {\color[HTML]{FF0000} \textbf{0.368}} & \textbf{0.388}                        & \textbf{0.374}                        & {\color[HTML]{FF0000} \textbf{0.387}} & 0.387                                 & 0.402          & 0.381              & 0.398     & 0.394       & 0.409               & 0.395                                 & 0.407 & 0.376              & 0.397    & 0.396        & 0.410        \\
\multicolumn{1}{c|}{}                          & 192 & {\color[HTML]{FF0000} \textbf{0.419}} & \textbf{0.417}                        & 0.428                                 & {\color[HTML]{FF0000} \textbf{0.414}} & 0.443                                 & 0.432          & 0.441              & 0.430     & 0.448       & 0.441               & 0.490                                 & 0.477 & \textbf{0.426}     & 0.432    & 0.445        & 0.440        \\
\multicolumn{1}{c|}{}                          & 336 & {\color[HTML]{FF0000} \textbf{0.454}} & {\color[HTML]{FF0000} \textbf{0.434}} & \textbf{0.462}                        & \textbf{0.437}                        & 0.492                                 & 0.457          & 0.500              & 0.459     & 0.492       & 0.465               & 0.510                                 & 0.480 & 0.469              & 0.457    & 0.487        & 0.465        \\
\multicolumn{1}{c|}{}                          & 720 & {\color[HTML]{FF0000} \textbf{0.459}} & {\color[HTML]{FF0000} \textbf{0.456}} & \textbf{0.482}                        & \textbf{0.466}                        & 0.504                                 & 0.484          & 0.552              & 0.507     & 0.521       & 0.504               & 0.568                                 & 0.538 & 0.518              & 0.504    & 0.512        & 0.510        \\
\multicolumn{1}{c|}{\multirow{-5}{*}{ETT h1}}  & avg & {\color[HTML]{FF0000} \textbf{0.425}} & {\color[HTML]{FF0000} \textbf{0.423}} & \textbf{0.436}                        & \textbf{0.426}                        & 0.456                                 & 0.443          & 0.468              & 0.448     & 0.463       & 0.454               & 0.490                                 & 0.475 & 0.447              & 0.447    & 0.460        & 0.456        \\ \midrule
\multicolumn{1}{c|}{}                          & 96  & {\color[HTML]{FF0000} \textbf{0.277}} & {\color[HTML]{FF0000} \textbf{0.327}} & \textbf{0.277}                        & \textbf{0.330}                        & 0.296                                 & 0.344          & 0.286              & 0.339     & 0.300       & 0.349               & 0.332                                 & 0.387 & 0.308              & 0.359    & 0.341        & 0.395        \\
\multicolumn{1}{c|}{}                          & 192 & {\color[HTML]{FF0000} \textbf{0.348}} & {\color[HTML]{FF0000} \textbf{0.375}} & \textbf{0.351}                        & \textbf{0.377}                        & 0.381                                 & 0.395          & 0.391              & 0.404     & 0.381       & 0.399               & 0.451                                 & 0.457 & 0.380              & 0.406    & 0.481        & 0.479        \\
\multicolumn{1}{c|}{}                          & 336 & \textbf{0.394}                        & \textbf{0.409}                        & {\color[HTML]{FF0000} \textbf{0.363}} & {\color[HTML]{FF0000} \textbf{0.394}} & 0.429                                 & 0.433          & 0.421              & 0.432     & 0.423       & 0.432               & 0.466                                 & 0.473 & 0.412              & 0.429    & 0.592        & 0.542        \\
\multicolumn{1}{c|}{}                          & 720 & \textbf{0.410}                        & \textbf{0.432}                        & {\color[HTML]{FF0000} \textbf{0.405}} & {\color[HTML]{FF0000} \textbf{0.427}} & 0.435                                 & 0.444          & 0.468              & 0.468     & 0.426       & 0.445               & 0.485                                 & 0.471 & 0.435              & 0.456    & 0.840        & 0.661        \\
\multicolumn{1}{c|}{\multirow{-5}{*}{ETT h2}}  & avg & \textbf{0.357}                        & \textbf{0.385}                        & {\color[HTML]{FF0000} \textbf{0.349}} & {\color[HTML]{FF0000} \textbf{0.382}} & 0.385                                 & 0.404          & 0.391              & 0.410     & 0.382       & 0.406               & 0.433                                 & 0.447 & 0.383              & 0.412    & 0.563        & 0.519        \\ \midrule
\multicolumn{1}{c|}{}                          & 96  & {\color[HTML]{FF0000} \textbf{0.304}} & {\color[HTML]{FF0000} \textbf{0.344}} & 0.334                                 & 0.368                                 & \textbf{0.322}                        & \textbf{0.363} & 0.327              & 0.364     & 0.341       & 0.376               & 0.337                                 & 0.374 & 0.323              & 0.364    & 0.345        & 0.373        \\
\multicolumn{1}{c|}{}                          & 192 & {\color[HTML]{FF0000} \textbf{0.355}} & {\color[HTML]{FF0000} \textbf{0.373}} & \textbf{0.358}                        & \textbf{0.375}                        & 0.361                                 & 0.385          & 0.367              & 0.386     & 0.380       & 0.394               & 0.382                                 & 0.398 & 0.371              & 0.391    & 0.381        & 0.391        \\
\multicolumn{1}{c|}{}                          & 336 & \textbf{0.385}                        & {\color[HTML]{FF0000} \textbf{0.395}} & {\color[HTML]{FF0000} \textbf{0.384}} & \textbf{0.397}                        & 0.401                                 & 0.414          & 0.393              & 0.403     & 0.419       & 0.418               & 0.420                                 & 0.423 & 0.398              & 0.408    & 0.415        & 0.415        \\
\multicolumn{1}{c|}{}                          & 720 & {\color[HTML]{FF0000} \textbf{0.443}} & {\color[HTML]{FF0000} \textbf{0.431}} & 0.456                                 & \textbf{0.435}                        & 0.452                                 & 0.443          & \textbf{0.451}     & 0.442     & 0.486       & 0.455               & 0.490                                 & 0.471 & 0.457              & 0.444    & 0.472        & 0.450        \\
\multicolumn{1}{c|}{\multirow{-5}{*}{ETT m1}}  & avg & {\color[HTML]{FF0000} \textbf{0.371}} & {\color[HTML]{FF0000} \textbf{0.385}} & \textbf{0.383}                        & \textbf{0.393}                        & 0.384                                 & 0.401          & 0.384              & 0.398     & 0.406       & 0.410               & 0.407                                 & 0.416 & 0.387              & 0.401    & 0.403        & 0.407        \\ \midrule
\multicolumn{1}{c|}{}                          & 96  & {\color[HTML]{FF0000} \textbf{0.161}} & {\color[HTML]{FF0000} \textbf{0.240}} & \textbf{0.170}                        & \textbf{0.251}                        & 0.177                                 & 0.263          & 0.174              & 0.257     & 0.183       & 0.266               & 0.186                                 & 0.275 & 0.184              & 0.267    & 0.193        & 0.292        \\
\multicolumn{1}{c|}{}                          & 192 & {\color[HTML]{FF0000} \textbf{0.224}} & {\color[HTML]{FF0000} \textbf{0.283}} & \textbf{0.235}                        & \textbf{0.295}                        & 0.248                                 & 0.311          & 0.236              & 0.299     & 0.252       & 0.312               & 0.259                                 & 0.323 & 0.246              & 0.304    & 0.284        & 0.361        \\
\multicolumn{1}{c|}{}                          & 336 & {\color[HTML]{FF0000} \textbf{0.284}} & {\color[HTML]{FF0000} \textbf{0.322}} & \textbf{0.300}                        & \textbf{0.336}                        & 0.313                                 & 0.353          & 0.301              & 0.339     & 0.314       & 0.351               & 0.349                                 & 0.386 & 0.311              & 0.348    & 0.384        & 0.429        \\
\multicolumn{1}{c|}{}                          & 720 & {\color[HTML]{FF0000} \textbf{0.382}} & {\color[HTML]{FF0000} \textbf{0.381}} & \textbf{0.391}                        & \textbf{0.392}                        & 0.418                                 & 0.415          & 0.400              & 0.400     & 0.411       & 0.406               & 0.559                                 & 0.511 & 0.418              & 0.414    & 0.556        & 0.523        \\
\multicolumn{1}{c|}{\multirow{-5}{*}{ETT m2}}  & avg & {\color[HTML]{FF0000} \textbf{0.262}} & {\color[HTML]{FF0000} \textbf{0.306}} & \textbf{0.274}                        & \textbf{0.318}                        & 0.289                                 & 0.335          & 0.277              & 0.323     & 0.290       & 0.333               & 0.338                                 & 0.373 & 0.289              & 0.333    & 0.354        & 0.401        \\ \midrule
\multicolumn{1}{c|}{}                          & 96  & {\color[HTML]{FF0000} \textbf{0.157}} & {\color[HTML]{FF0000} \textbf{0.202}} & 0.163                                 & \textbf{0.205}                        & 0.165                                 & 0.213          & \textbf{0.161}     & 0.208     & 0.175       & 0.215               & 0.171                                 & 0.227 & 0.175              & 0.217    & 0.196        & 0.256        \\
\multicolumn{1}{c|}{}                          & 192 & {\color[HTML]{FF0000} \textbf{0.202}} & {\color[HTML]{FF0000} \textbf{0.242}} & 0.207                                 & \textbf{0.245}                        & 0.214                                 & 0.255          & \textbf{0.207}     & 0.251     & 0.225       & 0.257               & 0.218                                 & 0.280 & 0.220              & 0.255    & 0.238        & 0.299        \\
\multicolumn{1}{c|}{}                          & 336 & {\color[HTML]{FF0000} \textbf{0.260}} & {\color[HTML]{FF0000} \textbf{0.286}} & 0.267                                 & \textbf{0.291}                        & 0.273                                 & 0.297          & \textbf{0.264}     & 0.293     & 0.279       & 0.298               & 0.265                                 & 0.317 & 0.279              & 0.297    & 0.281        & 0.330        \\
\multicolumn{1}{c|}{}                          & 720 & 0.341                                 & \textbf{0.339}                        & \textbf{0.338}                        & {\color[HTML]{FF0000} \textbf{0.337}} & 0.353                                 & 0.352          & 0.345              & 0.345     & 0.361       & 0.350               & {\color[HTML]{FF0000} \textbf{0.326}} & 0.351 & 0.356              & 0.348    & 0.345        & 0.381        \\
\multicolumn{1}{c|}{\multirow{-5}{*}{Weather}} & avg & {\color[HTML]{FF0000} \textbf{0.240}} & {\color[HTML]{FF0000} \textbf{0.267}} & \textbf{0.243}                        & \textbf{0.269}                        & 0.251                                 & 0.279          & 0.244              & 0.274     & 0.260       & 0.280               & 0.245                                 & 0.293 & 0.257              & 0.279    & 0.265        & 0.316        \\ \midrule
\multicolumn{1}{c|}{}                          & 96  & {\color[HTML]{FF0000} \textbf{0.135}} & {\color[HTML]{FF0000} \textbf{0.227}} & 0.166                                 & 0.260                                 & \textbf{0.147}                        & 0.245          & 0.156              & 0.247     & 0.148       & \textbf{0.240}      & 0.171                                 & 0.260 & 0.180              & 0.272    & 0.210        & 0.301        \\
\multicolumn{1}{c|}{}                          & 192 & {\color[HTML]{FF0000} \textbf{0.152}} & {\color[HTML]{FF0000} \textbf{0.242}} & 0.175                                 & 0.261                                 & \textbf{0.161}                        & 0.257          & 0.170              & 0.260     & 0.164       & \textbf{0.256}      & 0.177                                 & 0.268 & 0.187              & 0.279    & 0.210        & 0.304        \\
\multicolumn{1}{c|}{}                          & 336 & {\color[HTML]{FF0000} \textbf{0.170}} & {\color[HTML]{FF0000} \textbf{0.260}} & 0.193                                 & 0.282                                 & \textbf{0.176}                        & 0.274          & 0.187              & 0.278     & 0.177       & \textbf{0.270}      & 0.190                                 & 0.284 & 0.204              & 0.295    & 0.223        & 0.319        \\
\multicolumn{1}{c|}{}                          & 720 & {\color[HTML]{FF0000} \textbf{0.207}} & {\color[HTML]{FF0000} \textbf{0.293}} & 0.233                                 & 0.314                                 & {\color[HTML]{FF0000} \textbf{0.207}} & \textbf{0.304} & 0.227              & 0.312     & 0.228       & 0.313               & 0.228                                 & 0.316 & 0.245              & 0.328    & 0.257        & 0.349        \\
\multicolumn{1}{c|}{\multirow{-5}{*}{Electricity}}     & avg & {\color[HTML]{FF0000} \textbf{0.166}} & {\color[HTML]{FF0000} \textbf{0.255}} & 0.191                                 & 0.279                                 & \textbf{0.172}                        & 0.270          & 0.185              & 0.274     & 0.179       & \textbf{0.269}      & 0.191                                 & 0.282 & 0.204              & 0.293    & 0.225        & 0.318        \\ \midrule
\multicolumn{2}{c|}{Total AVG}                       & {\color[HTML]{FF0000} \textbf{0.303}} & {\color[HTML]{FF0000} \textbf{0.336}} & \textbf{0.312}                        & \textbf{0.344}                        & 0.322                                 & 0.355          & 0.324              & 0.354     & 0.330       & 0.358               & 0.350                                 & 0.381 & 0.327              & 0.360    & 0.378        & 0.402        \\ \midrule
\multicolumn{2}{c|}{1st Times}                       & {\color[HTML]{FF0000} \textbf{26}}    & {\color[HTML]{FF0000} \textbf{25}}    & \textbf{4}                            & \textbf{6}                            & 1                                     & 0              & 0                  & 0         & 0           & 0                   & 1                                     & 0     & 0                  & 0        & 0            & 0            \\ \midrule
\end{tabular}
    }
  \end{table*}
\subsection{Experiment setting}
% \usepackage{multirow}
% \usepackage{graphicx}

\textbf{Experimental Datasets:} To validate the forecasting accuracy of our mode, we selected six commonly used real-world datasets in LTSF (ETTh1, ETTh2, ETTm1, ETTm2, Weather, Electricity) for experimentation.

\textbf{Baseline Models:} Based on recency, innovation and forecasting performance, we selected seven well-regarded time series models in the field as our baselines. These include both Linear-based and Transformer-based models: (1) WPMixer (2) SDE (3) TimeMixer (4) iTransformer (5) FreTS (6) PatchTST and (7) DLinear.

\textbf{Evaluation Metrics:} This experiment uses Mean Squared Error (MSE) and Mean Absolute Error (MAE) as the evaluation metrics for the models.

\textbf{Experimental Setup:} All experiments were implemented with PyTorch on a single NVIDIA 3090 24GB GPU and baseline model metrics were derived from local testing via their original codebase scripts.

\subsection{Main results}

All results in this experiment were obtained by running the original code scripts locally(except for FreTS) and the complete results are shown in Table 1. The results clearly show that our model's performance is significantly better than the baseline models. Specifically, compared to the best results from other models, our model achieves an average reduction in MSE of 2.5\% on ETTh1, 3.1\% on ETTm1, 4.3\% on ETTm2, 1.2\% on Weather, and 3.4\% on Electricity. Similarly, for these datasets, the MAE is reduced by 0.7\%, 2.0\%, 2.8\%, 0.7\% and 5.2\% respectively. Although our model does not achieve the best prediction performance on every dataset, McWC's accuracy remains very close to the top-performing models in those cases. Furthermore, in terms of the number of top rankings achieved, our proposed McWC obtained 51 best results and 9 second-best results, far surpassing all other models in the experiment. This demonstrates that McWC possesses accurate and versatile forecasting capabilities for the vast majority of natural time series.

\subsection{Ablation experiment}

\begin{figure*}[!t]  % [!t] 表示优先放在页面顶部，可选参数
	\centering
	% 调整宽度为单栏合适尺寸（双栏文档中\textwidth此时是整页宽度）
	\includegraphics[width=0.73\textwidth]{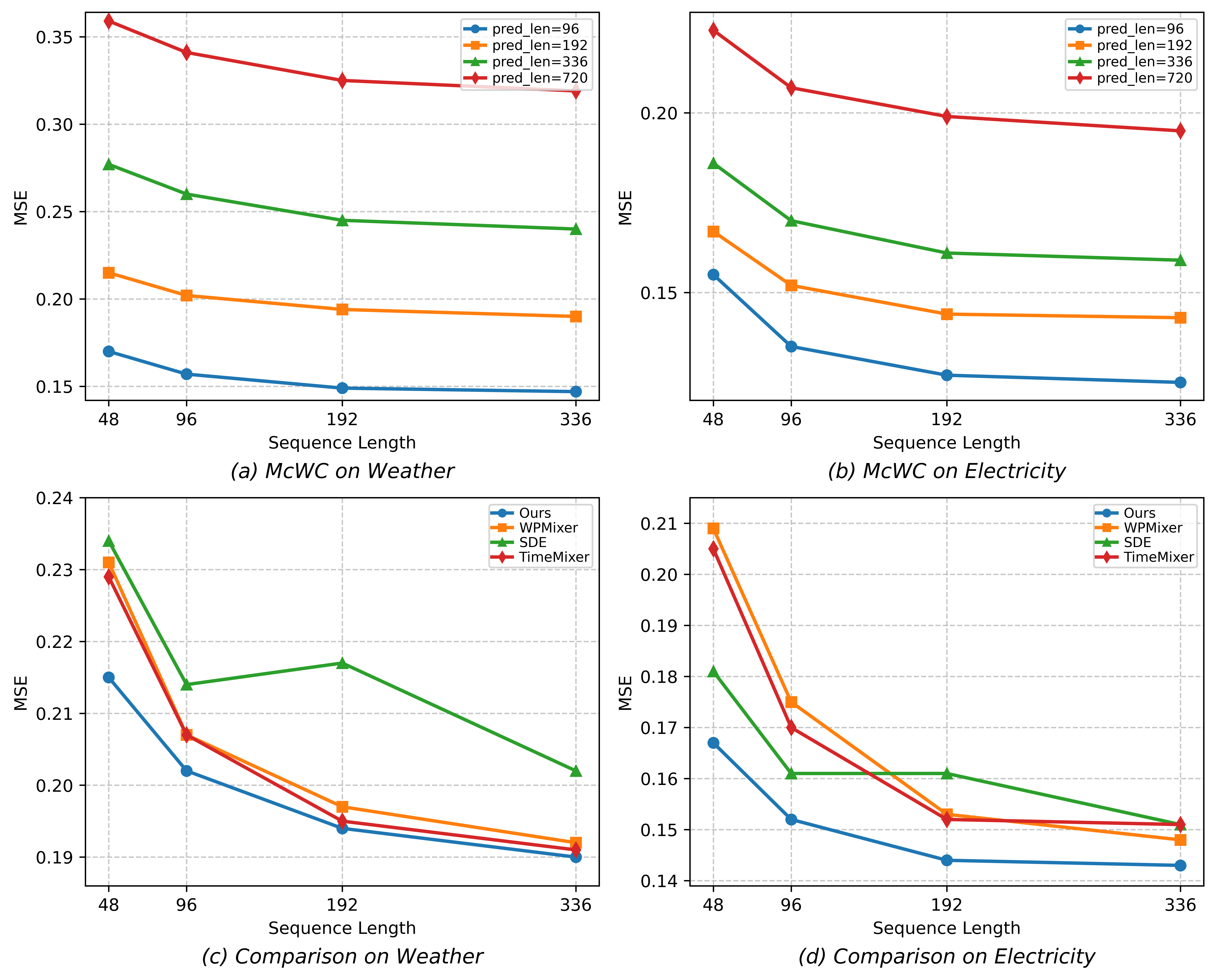}
	% 标题内容保持不变
	\caption{Experiments of the information extraction performance by extending the sequence length on datasets(a) Weather and (b) Electricity. Comparison of forecasting performance between ours and baseline models by varying look-back windows on datasets(c) Weather and (d) Electricity.}
	\label{FIG:2}
\end{figure*}

\begin{table}[!t]
  \caption{Ablation study of McWC: TL means train loss; \Checkmark means module included in backbone, \XSolid means not included. Best results in \color[HTML]{FF0000} \textbf{bold red}.}
  \label{tab:ablation}
  \setlength{\tabcolsep}{4pt} % 稍微增加列间距以获得更好的外观
  \footnotesize
  % 使用 booktabs 风格并为前三列设置固定的居中列宽
  % 请确保在 LaTeX 导言区添加了 \usepackage{booktabs} 和 \usepackage{array}
  \begin{tabular}{>{\centering\arraybackslash}p{3.5em} >{\centering\arraybackslash}p{3.5em} >{\centering\arraybackslash}p{3.5em} c c c c}
    \toprule
    \multicolumn{3}{c}{Modules} & ETTh1 & ETTm2 & Weather & Electricity \\ 
    \cmidrule(r){1-3} \cmidrule(l){4-7} % 使用 cmidrule 创建分段的横线
    CEB & McB & TL & MSE & MSE & MSE & MSE \\ 
    \midrule
    \Checkmark & \Checkmark & \Checkmark & {\color[HTML]{FF0000} \textbf{0.425}} & {\color[HTML]{FF0000} \textbf{0.262}} & {\color[HTML]{FF0000} \textbf{0.240}} & {\color[HTML]{FF0000} \textbf{0.1660}} \\
    \XSolid & \Checkmark & \Checkmark & {\color[HTML]{FF0000} \textbf{0.425}} & 0.263 & 0.244 & 0.167 \\
    \Checkmark & \XSolid & \Checkmark & 0.434 & 0.274 & 0.245 & 0.190 \\
    \Checkmark & \Checkmark & \XSolid & 0.443 & 0.265 & 0.241 & 0.1667 \\
    \Checkmark & \XSolid & \XSolid & 0.447 & 0.274 & 0.244 & 0.192 \\
    \XSolid & \Checkmark & \XSolid & 0.443 & 0.267 & 0.242 & 0.1667 \\
    \XSolid & \XSolid & \Checkmark & 0.434 & 0.278 & 0.261 & 0.201 \\ 
    \midrule
    \multicolumn{3}{c}{Backbone (without modules)} & 0.451 & 0.280 & 0.260 & 0.200 \\ 
    \bottomrule
  \end{tabular}
\end{table}

\begin{table}[!t]
\centering
  \caption{Comparison of computational cost.The best results are highlighted in \color[HTML]{FF0000} \textbf{bold red.} }
  \label{tab:ablation}
  \setlength{\tabcolsep}{3pt}
  \footnotesize
  \begin{tabular}{cc|clclclcl}
\midrule
\multicolumn{2}{c|}{\multirow{2}{*}{Models}} &
  \multicolumn{2}{c}{McWC} &
  \multicolumn{2}{c}{WPMixer} &
  \multicolumn{2}{c}{SDE} &
  \multicolumn{2}{c}{TimeMixer} \\
\multicolumn{2}{c|}{}       & \multicolumn{2}{c}{2025}          & \multicolumn{2}{c}{2025}   & \multicolumn{2}{c}{2025}   & \multicolumn{2}{c}{2024}   \\ \midrule
\multicolumn{2}{c|}{Metric} & \multicolumn{2}{c}{GFLOPs}        & \multicolumn{2}{c}{GFLOPs} & \multicolumn{2}{c}{GFLOPs} & \multicolumn{2}{c}{GFLOPs} \\ \midrule
\multicolumn{1}{c|}{\multirow{5}{*}{ETT m2}} &
  96 &
  \multicolumn{2}{c}{\color[HTML]{FF0000}\textbf{0.18}} &
  \multicolumn{2}{c}{14.24} &
  \multicolumn{2}{c}{2.47} &
  \multicolumn{2}{c}{2.58} \\
\multicolumn{1}{c|}{} & 192 & \multicolumn{2}{c}{\color[HTML]{FF0000}\textbf{0.26}} & \multicolumn{2}{c}{12.57}  & \multicolumn{2}{c}{2.59}   & \multicolumn{2}{c}{3.06}   \\
\multicolumn{1}{c|}{} & 336 & \multicolumn{2}{c}{\color[HTML]{FF0000}\textbf{0.38}} & \multicolumn{2}{c}{9.20}   & \multicolumn{2}{c}{2.78}   & \multicolumn{2}{c}{3.76}   \\
\multicolumn{1}{c|}{} & 720 & \multicolumn{2}{c}{\color[HTML]{FF0000}\textbf{0.71}} & \multicolumn{2}{c}{14.78}  & \multicolumn{2}{c}{3.27}   & \multicolumn{2}{c}{5.65}   \\
\multicolumn{1}{c|}{} & avg & \multicolumn{2}{c}{\color[HTML]{FF0000}\textbf{0.38}} & \multicolumn{2}{c}{12.69}  & \multicolumn{2}{c}{2.77}   & \multicolumn{2}{c}{3.760}  \\ \midrule
\multicolumn{1}{c|}{\multirow{5}{*}{Weather}} &
  96 &
  \multicolumn{2}{c}{\color[HTML]{FF0000}\textbf{0.53}} &
  \multicolumn{2}{c}{14.03} &
  \multicolumn{2}{c}{37.62} &
  \multicolumn{2}{c}{5.43} \\
\multicolumn{1}{c|}{} & 192 & \multicolumn{2}{c}{\color[HTML]{FF0000}\textbf{0.77}} & \multicolumn{2}{c}{7.29}   & \multicolumn{2}{c}{38.00}  & \multicolumn{2}{c}{6.14}   \\
\multicolumn{1}{c|}{} & 336 & \multicolumn{2}{c}{\color[HTML]{FF0000}\textbf{1.14}} & \multicolumn{2}{c}{3.73}   & \multicolumn{2}{c}{38.55}  & \multicolumn{2}{c}{7.20}   \\
\multicolumn{1}{c|}{} & 720 & \multicolumn{2}{c}{\color[HTML]{FF0000}\textbf{2.13}} & \multicolumn{2}{c}{12.02}  & \multicolumn{2}{c}{160.10} & \multicolumn{2}{c}{10.03}  \\
\multicolumn{1}{c|}{} & avg & \multicolumn{2}{c}{\color[HTML]{FF0000}\textbf{1.14}} & \multicolumn{2}{c}{9.26}   & \multicolumn{2}{c}{68.56}  & \multicolumn{2}{c}{7.20}   \\ \midrule
\end{tabular}
\end{table}

To demonstrate the effectiveness of the modules we designed and added, we conducted ablation studies on four datasets: ETTh1, ETTm2, Weather, and Electricity. This experiment consists of configurations with different module combinations and the specific modules included in each case are detailed in Table 2. The experimental results clearly show that the McB module, which constructs periodic patterns, enhances performance on most datasets. The TL and CEB modules function by addressing intra-channel autocorrelation and inter-channel correlation in different dataset respectively. Additionally, we compared the Backbone against an MLP by conducting forecasting across all datasets to validate the effectiveness of the Backbone. Their average MSE and MAE were 0.327, 0.355 and 0.333, 0.359 respectively. Compared to the MLP, our Backbone achieved an improvement of 1.8\% in average MSE and 1.1\% in average MAE across all datasets, demonstrating its effectiveness. Therefore, through the combined effect of our designed and introduced modules, McWC achieves outstanding performance on the majority of datasets. This sufficiently proves the importance and necessity of the modules we proposed and incorporated into McWC.

\subsection{Model Efficiency}
To evaluate the efficiency of our proposed McWC, we measured its computational cost in Giga Floating Point Operations (GFLOPs) under different testing conditions across various datasets. For the experiments, we calculated GFLOPs using the parameters that yielded the optimal results for each model under its respective test conditions. As shown in Table 3, McWC requires nearly an order of magnitude less computational resources than the other three baseline models across all tested scenarios. This demonstrates that our proposed McWC model achieves stable and highly efficient operational performance while maintaining forecasting accuracy.

\subsection{Varing look-back window}

It is evident that the length of the look-back window is directly proportional to the amount of historical information it contains. However, without strong information extraction and integration capabilities, excessive historical data may lead to overfitting and adversely affect forecast performance. As a deep learning model that explicitly models the intrinsic components of sequential data, our proposed McWC possesses powerful capabilities for extracting historical information. To validate this, we conducted experiments on the Weather and Electricity datasets with varying look-back window lengths: look-back length $=\left\{48,96,192,336\right\}$ and forecast length $=\left\{96,192,336,720\right\}$. As shown in Figure 2(a) and 2(b), the forecasting quality of McWC improves consistently as the look-back window increases. Furthermore, we compared the information extraction capability of McWC against multiple baseline models by setting a fixed prediction length of 192 and testing different look-back window lengths. The results in Figure 2(c) and 2(d) demonstrate that McWC's prediction quality is positively correlated with the look-back window length and its information extraction capability outperforms the baseline models.

\section{Conclusion}
In this paper, we propose McWC, a computationally efficient model for long-term time series forecasting. Based on the idea of component decomposition, McWC adopts the McB, CEB, and MWB modules to model periodicity, trend and channel correlation, improving its ability to capture complex patterns and data mutations. We further introduce FreDF to optimize the loss function, enhancing intra-channel autocorrelation modeling and prediction performance. Extensive experiments on real-world datasets show that McWC effectively exploits historical information and achieves high-quality forecasting.

\section*{ACKNOWLEDGMENT}
This work was supported by the National Natural Science Foundation of China (Grant No. 82574111).


%%%%%%%%%%%%%%%%%%%%%%%%%%%%%%%%%%%%%%%%%%%%%%%%%%%%%%%%%%%%%%%%%%%%%%%%%%%%%%%%
%%%%%%%%%%%%%%%%%%%%%%%%%%%%%%%%%%%%%%%%%%%%%%%%%%%%%%%%%%%%%%%%%%%%%%%%%%%%%%%%
\bibliographystyle{IEEEtran}
\bibliography{McWC}

@article{zhang_financial_2023,
	title = {Financial time series forecasting based on momentum-driven graph signal processing},
	volume = {53},
	issn = {1573-7497},
	doi = {10.1007/s10489-023-04563-y},
	number = {18},
	journal = {Applied Intelligence},
	author = {Zhang, Shengen and Ma, Xu and Fang, Zhen and Pan, Huifeng and Yang, Guangbing and Arce, Gonzalo R.},
	month = sep,
	year = {2023},
	pages = {20950--20966},
}

@article{hadry_telling_2025,
	title = {Telling fortunes? {Evaluation} of traffic forecasting models using traffic and context features},
	volume = {55},
	issn = {1573-7497},
	doi = {10.1007/s10489-025-06565-4},
	number = {10},
	journal = {Applied Intelligence},
	author = {Hadry, Marius and Bauer, André and Leppich, Robert and Lesch, Veronika and Kounev, Samuel},
	month = jun,
	year = {2025},
	pages = {755},
}

@article{hu_decomposition_2025,
	title = {Decomposition dynamic multi-graph convolutional recurrent network for traffic forecasting},
	volume = {55},
	issn = {1573-7497},
	doi = {10.1007/s10489-025-06503-4},
	number = {7},
	journal = {Applied Intelligence},
	author = {Hu, Longfei and Wei, Lai and Lin, Yeqing},
	month = mar,
	year = {2025},
	pages = {595},
}

@article{zhang2024sageformer,
  title={SageFormer: Series-aware framework for long-term multivariate time-series forecasting},
  author={Zhang, Zhenwei and Meng, Linghang and Gu, Yuantao},
  journal={IEEE Internet of Things Journal},
  volume={11},
  number={10},
  pages={18435--18448},
  year={2024},
  publisher={IEEE}
}

@inproceedings{wangcard,
  title={CARD: Channel Aligned Robust Blend Transformer for Time Series Forecasting},
  author={Wang, Xue and Zhou, Tian and Wen, Qingsong and Gao, Jinyang and Ding, Bolin and Jin, Rong},
  booktitle={The Twelfth International Conference on Learning Representations}
}

@inproceedings{weng2025sde,
  title={SDE: A Simplified and Disentangled Dependency Encoding Framework for State Space Models in Time Series Forecasting},
  author={Weng, Zixuan and Han, Jindong and Jiang, Wenzhao and Liu, Hao},
  booktitle={Proceedings of the 31st ACM SIGKDD Conference on Knowledge Discovery and Data Mining V. 2},
  pages={3168--3179},
  year={2025}
}

@article{lecun2002gradient,
  title={Gradient-based learning applied to document recognition},
  author={LeCun, Yann and Bottou, L{\'e}on and Bengio, Yoshua and Haffner, Patrick},
  journal={Proceedings of the IEEE},
  volume={86},
  number={11},
  pages={2278--2324},
  year={2002},
  publisher={Ieee}
}

@inproceedings{zhou2022fedformer,
  title={Fedformer: Frequency enhanced decomposed transformer for long-term series forecasting},
  author={Zhou, Tian and Ma, Ziqing and Wen, Qingsong and Wang, Xue and Sun, Liang and Jin, Rong},
  booktitle={International conference on machine learning},
  pages={27268--27286},
  year={2022},
  organization={PMLR}
}

@inproceedings{murad2025wpmixer,
  title={Wpmixer: Efficient multi-resolution mixing for long-term time series forecasting},
  author={Murad, Md Mahmuddun Nabi and Aktukmak, Mehmet and Yilmaz, Yasin},
  booktitle={Proceedings of the AAAI Conference on Artificial Intelligence},
  volume={39},
  number={18},
  pages={19581--19588},
  year={2025}
}

@article{qin2026forecasting,
  title={Forecasting short-term wind power with multi-view attention mechanism and dual recurrent neural networks},
  author={Qin, Chaoyong and Xie, Jialin and Cao, Yun and Zhu, Bangzhu},
  journal={Expert Systems with Applications},
  volume={297},
  pages={129472},
  year={2026},
  publisher={Elsevier}
}

@article{vaswani_attention_2017,
	title = {Attention is all you need},
	volume = {30},
	journal = {Advances in neural information processing systems},
	author = {Vaswani, Ashish and Shazeer, Noam and Parmar, Niki and Uszkoreit, Jakob and Jones, Llion and Gomez, Aidan N and Kaiser, Lukasz and Polosukhin, Illia},
	year = {2017},
}

@inproceedings{huang_timekan_2025,
	title = {{TimeKAN}: {KAN}-based {Frequency} {Decomposition} {Learning} {Architecture} for {Long}-term {Time} {Series} {Forecasting}},
	language = {en},
	booktitle = {The {Thirteenth} {International} {Conference} on {Learning} {Representations}},
	author = {Huang, Songtao and Zhao, Zhen and Li, Can and Bai, Lei},
	year = {2025},
	pages = {93540--93555},
}

@inproceedings{wang_fredf_2024,
	title = {Fredf: {Learning} to forecast in frequency domain},
	language = {en},
	booktitle = {The {Thirteenth} {International} {Conference} on {Learning} {Representations}},
	author = {Wang, Hao and Pan, Licheng and Chen, Zhichao and Yang, Degui and Zhang, Sen and Yang, Yifei and Liu, Xinggao and Li, Haoxuan and Tao, Dacheng},
	year = {2024},
	pages = {7329--7358},
}

@article{liang_crossformer_2024,
	title = {{CrossFormer}: {Cross}-{Modal} {Representation} {Learning} via {Heterogeneous} {Graph} {Transformer}},
	volume = {20},
	number = {12},
	journal = {ACM Trans. Multim. Comput. Commun. Appl.},
	author = {Liang, Xiao and Yang, Erkun and Deng, Cheng and Yang, Yanhua},
	month = dec,
	year = {2024},
	pages = {380:1--380:21},
}

@inproceedings{wang_timemixer_2024,
	title = {{TimeMixer}: {Decomposable} {Multiscale} {Mixing} for {Time} {Series} {Forecasting}},
	language = {en},
	booktitle = {{ICLR}},
	author = {Wang, Shiyu and Wu, Haixu and Shi, Xiaoming and Hu, Tengge and Luo, Huakun and Ma, Lintao and Zhang, James Y. and Zhou, Jun},
	year = {2024},
	pages = {4166--4192},

}

@inproceedings{wu_autoformer_2021,
	title = {Autoformer: {Decomposition} {Transformers} with {Auto}-{Correlation} for {Long}-{Term} {Series} {Forecasting}},
	volume = {34},
	language = {en},
	booktitle = {Advances in {Neural} {Information} {Processing} {Systems}},
	author = {Wu, Haixu and Xu, Jiehui and Wang, Jianmin and Long, Mingsheng},
	year = {2021},
	pages = {22419--22430},
}

@inproceedings{liu_itransformer_2024,
	title = {{iTransformer}: {Inverted} {Transformers} {Are} {Effective} for {Time} {Series} {Forecasting}},
	language = {en},
	booktitle = {The {Twelfth} {International} {Conference} on {Learning} {Representations}},
	author = {Liu, Yong and Hu, Tengge and Zhang, Haoran and Wu, Haixu and Wang, Shiyu and Ma, Lintao and Long, Mingsheng},
	year = {2024},
	pages = {4004--4028},
}

@inproceedings{nie_time_2023,
	title = {A {Time} {Series} is {Worth} 64 {Words}: {Long}-term {Forecasting} with {Transformers}},
	language = {en},
	booktitle = {The {Eleventh} {International} {Conference} on {Learning} {Representations}},
	author = {Nie, Yuqi and Nguyen, Nam H. and Sinthong, Phanwadee and Kalagnanam, Jayant},
	year = {2023},
	pages = {33132--33155},
}

@inproceedings{wang_micn_2023,
	title = {{MICN}: {Multi}-scale {Local} and {Global} {Context} {Modeling} for {Long}-term {Series} {Forecasting}},
	language = {en},
	booktitle = {The eleventh international conference on learning representations},
	author = {Wang, Huiqiang and Peng, Jian and Huang, Feihu and Wang, Jince and Chen, Junhui and Xiao, Yifei},
	year = {2023},
	pages = {13014--13035},
}

@inproceedings{wu_timesnet_2023,
  title = {{TimesNet}: {Temporal} {2D}-{Variation} {Modeling} for {General} {Time} {Series} {Analysis}},
	language = {en},
	booktitle = {The {Eleventh} {International} {Conference} on {Learning} {Representations}},
	author = {Wu, Haixu and Hu, Tengge and Liu, Yong and Zhou, Hang and Wang, Jianmin and Long, Mingsheng},
	year = {2023},
	pages = {6423--6445},
}

@inproceedings{zeng_are_2023,
	title = {Are transformers effective for time series forecasting?},
	volume = {37},
	booktitle = {Proceedings of the {AAAI} conference on artificial intelligence},
	author = {Zeng, Ailing and Chen, Muxi and Zhang, Lei and Xu, Qiang},
	year = {2023},
	note = {Issue: 9},
	pages = {11121--11128},
}

@article{yi_frequency-domain_2023,
	title = {Frequency-domain mlps are more effective learners in time series forecasting},
	volume = {36},
	journal = {Advances in Neural Information Processing Systems},
	author = {Yi, Kun and Zhang, Qi and Fan, Wei and Wang, Shoujin and Wang, Pengyang and He, Hui and An, Ning and Lian, Defu and Cao, Longbing and Niu, Zhendong},
	year = {2023},
	pages = {76656--76679},
}
%%%%%%%%%%%%%%%%%%%%%%%%%%%%%%%%%%%%%%%%%%%%%%%%%%%%%%%%%%%%%%%%%%%%%%%%%%%%%%%%
\end{document}